\documentclass[11pt,a4paper]{article}
\usepackage[hyperref]{emnlp2020}
\usepackage{times}
\usepackage{latexsym}

\usepackage{todonotes}

\usepackage{microtype}
\usepackage{xcolor}
\usepackage{url}
\usepackage{multirow}
\usepackage{amssymb}
\usepackage{bbm}
\usepackage{graphicx}
\usepackage{multicol}
\usepackage{makecell}
\usepackage{pifont}
\usepackage{array}
\newcolumntype{L}[1]{>{\raggedright\let\newline\\\arraybackslash\hspace{0pt}}m{#1}}
\newcolumntype{C}[1]{>{\centering\let\newline\\\arraybackslash\hspace{0pt}}m{#1}}
\newcolumntype{R}[1]{>{\raggedleft\let\newline\\\arraybackslash\hspace{0pt}}m{#1}}
\newcolumntype{B}[1]{>{\raggedright\let\newline\\\arraybackslash\hspace{0pt}}p{#1}}
\newcolumntype{N}[1]{>{\centering\let\newline\\\arraybackslash\hspace{0pt}}p{#1}}
\newcolumntype{M}[1]{>{\raggedleft\let\newline\\\arraybackslash\hspace{0pt}}p{#1}}
\aclfinalcopy % Uncomment this line for the final submission
\title{FIND: Human-in-the-Loop Debugging Deep Text Classifiers}

\author{Piyawat Lertvittayakumjorn,  Lucia Specia,  Francesca Toni\\
  Department of Computing,
  Imperial College London, UK\\
  \texttt{\{pl1515, l.specia, ft\}@imperial.ac.uk}}

\date{}

\begin{document}
\maketitle
\begin{abstract}
Since obtaining a perfect training dataset (i.e., a dataset which is considerably large, unbiased, and well-representative of unseen cases) is hardly possible, many real-world text classifiers are trained on the available, yet imperfect, datasets. 
These classifiers are thus likely to have undesirable properties. For instance, they may have biases against some sub-populations or may not work effectively in the wild due to overfitting. 
In this paper, we propose \textbf{FIND} -- a framework which enables humans to debug deep learning text classifiers by disabling irrelevant hidden features.
Experiments show that by using FIND, humans can improve CNN text classifiers which were trained under different types of imperfect datasets (including datasets with biases and datasets with dissimilar train-test distributions).
\end{abstract}

\section{Introduction}
Deep learning has become the dominant approach to address most Natural Language Processing (NLP) tasks, including text classification. With sufficient and high-quality training data, deep learning models can perform incredibly well \cite{zhang2015character,wang2019glue}. However, in real-world cases, such ideal datasets are scarce. Often times, the available datasets are small, 
full of regular but irrelevant words,
and contain unintended biases \cite{wiegand-etal-2019-detection,gururangan-etal-2018-annotation}. These can lead to suboptimal models with undesirable properties. For example, the models may have biases against some sub-populations or may not work effectively in the wild as they overfit the imperfect training data.

To improve the models, previous work has looked into different techniques beyond standard model fitting. If the weaknesses of the training datasets or the models are anticipated, strategies can be tailored to mitigate such weaknesses. For example, augmenting the training data with gender-swapped input texts helps reduce gender bias in the models \cite{park-etal-2018-reducing,zhao-etal-2018-gender}. Adversarial training can prevent the models from exploiting irrelevant and/or protected features \cite{jaiswal2019invariant,zhang2018mitigating}. With a limited number of training examples, using human rationales or prior knowledge together with training labels can help the models perform better \cite{zaidan-etal-2007-using,bao-etal-2018-deriving,liu-avci-2019-incorporating}.

Nonetheless, there are side-effects of sub-optimal datasets that cannot be predicted and are only found after training thanks to post-hoc error analysis. To rectify such problems, there have been attempts to enable humans to fix the trained models (i.e., to perform \textit{model debugging}) \cite{stumpf2009interacting,teso2019explanatory}. Since the models are usually too complex to understand, manually modifying the model parameters is not possible. Existing techniques, therefore, allow humans to provide feedback on individual predictions instead. Then, additional training examples are created based on the feedback to retrain the models. 
However, such local improvements for individual predictions could add up to inferior overall performance \cite{wu2019local}.
Furthermore, these existing techniques allow us to rectify only errors related to examples at hand but provide no way to fix problems kept hidden in the model parameters. 

In this paper, we propose a framework which allows humans to debug and improve deep text classifiers by disabling hidden features which are irrelevant to the classification task.
We name this framework \textbf{FIND} (\textbf{F}eature \textbf{I}nvestigation a\textbf{N}d \textbf{D}isabling).
FIND exploits an explanation method, namely layer-wise relevance propagation (LRP) \cite{arras-etal-2016-explaining}, to understand the behavior of a classifier when it predicts each training instance. Then it aggregates all the information using word clouds to create a global visual picture of the model.
This enables humans to comprehend the features automatically learned by the deep classifier and then decide to disable some features that could undermine the prediction accuracy during testing. The main differences between our work and existing work are: 
(i) first, FIND leverages human feedback on the model components, not the individual predictions, to perform debugging; 
(ii) second, FIND targets deep text classifiers which are more convoluted than traditional classifiers used in existing work (such as Naive Bayes classifiers and Support Vector Machines).

We conducted three human experiments (one feasibility study and two debugging experiments) to demonstrate the usefulness of FIND. 
For all the experiments, we used as classifiers convolutional neural networks (CNNs) \cite{Kim-2014-convolutional}, 
which are a popular, well-performing architecture for many text classification tasks including the tasks we experimented with \cite{gamback-sikdar-2017-using,johnson-zhang-2015-effective,zhang-etal-2019-integrating}.
The overall results show that FIND with human-in-the-loop can improve the text classifiers and mitigate the said problems in the datasets. 
After the experiments, we discuss the generalization of the proposed framework to other tasks and models.
Overall, the {\bf main contributions} of this paper are:
\begin{itemize}
    \setlength\itemsep{0em}
    \item We propose using word clouds as visual explanations of the features learned.
    \item We propose a technique to disable the learned features which are irrelevant or harmful to the classification task so as to improve the classifier. This technique and the word clouds form the human-debugging framework -- FIND. 
    \item We conduct three human experiments that demonstrate the effectiveness of FIND in different scenarios. The results not only highlight the usefulness of our approach but also reveal interesting behaviors of CNNs for text classification. 
\end{itemize}

The rest of this paper is organized as follows. 
Section \ref{sec:rel} explains related work about analyzing, explaining, and human-debugging text classifiers. 
Section \ref{sec:met} proposes FIND, our debugging framework. 
Section \ref{sec:exp} explains the experimental setup followed by the three human experiments in Section \ref{sec:exp1} to \ref{sec:exp3}.
Finally, Section \ref{sec:con} discusses generalization of the framework and concludes the paper.
Code and datasets of this paper are available at \url{https://github.com/plkumjorn/FIND}.

\section{Related Work} \label{sec:rel}
\textbf{Analyzing deep NLP models} -- 
There has been substantial work in gaining better understanding of complex, deep neural NLP models. 
By visualizing dense hidden vectors, \citet{li-etal-2016-visualizing} found that some dimensions of the final representation learned by recurrent neural networks capture the effect of intensification and negation in the input text. 
\citet{karpathy2015visualizing} revealed the existence of interpretable cells in a character-level LSTM model for language modelling. For example, they found a cell acting as a line length counter and cells checking if the current letter is inside a parenthesis or a quote. 
\citet{jacovi-etal-2018-understanding} presented interesting findings about CNNs for text classification including the fact that one convolutional filter may detect more than one n-gram pattern and may also suppress negative n-grams.
Many recent papers studied several types of knowledge in BERT \cite{devlin-etal-2019-bert}, a deep transformer-based model for language understanding, and found that syntactic information is mostly captured in the middle BERT layers while the final BERT layers are the most task-specific \cite{rogers2020primer}.
Inspired by many findings, we make the assumption that each dimension of the final representation (i.e., the vector before the output layer) captures patterns or qualities in the input which are useful for classification. Therefore, understanding the roles of these dimensions (we refer to them as \textit{features}) is a prerequisite for effective human-in-the-loop model debugging, and we exploit an explanation method to gain such an understanding.

\noindent \textbf{Explaining predictions from text classifiers} -- 
Several methods have been devised to generate explanations supporting  classifications in many forms, such as natural language texts \cite{liu-etal-2019-towards-explainable}, rules \cite{ribeiro2018anchors}, extracted rationales \cite{lei-etal-2016-rationalizing}, and attribution scores \cite{lertvittayakumjorn-toni-2019-human}. 
Some explanation methods, such as LIME \cite{ribeiro2016lime} and SHAP \cite{lundberg2017unified}, are model-agnostic and do not require access to model parameters. 
Other methods access the model architectures and parameters to generate the explanations, such as DeepLIFT \cite{shrikumar2017deeplift} and LRP (layer-wise relevance propagation) \cite{bach2015lrp,arras-etal-2016-explaining}. In this work, we use LRP to explain not the predictions but the learned features so as to expose the model behavior to humans and enable informed model debugging.

\noindent \textbf{Debugging text classifiers using human feedback} -- Early work in this area comes from the human-computer interaction community.
\citet{stumpf2009interacting} studied the types of feedback humans usually give in response to machine-generated predictions and explanations. Also, some of the feedback collected (i.e., important words of each category) was used to improve the classifier via a user co-training approach. 
\citet{kulesza2015principles} presented an explanatory debugging approach in which the system explains to users how it made each prediction, and the users then rectify the model by adding/removing words from the explanation and adjusting important weights. 
Even without explanations shown, an active learning framework proposed by \citet{settles-2011-closing} asks humans to iteratively label some chosen features (i.e., words) and adjusts the model parameters that correspond to the features. 
However, these early works target simpler machine learning classifiers (e.g., Naive Bayes classifiers with bag-of-words) and it is not clear how to apply the proposed approaches to deep text classifiers. 

Recently, there have been new attempts to use explanations and human feedback to debug classifiers in general. Some of them were tested on traditional text classifiers. 
For instance, \citet{ribeiro2016lime} showed a set of LIME explanations for individual SVM predictions to humans and asked them to remove irrelevant words from the training data in subsequent training. The process was run for three rounds to iteratively improve the classifiers.
\citet{teso2019explanatory} proposed CAIPI, which is an explanatory interactive learning framework. At each iteration, it selects an unlabelled example to predict and explain to users using LIME, and the users respond by removing irrelevant features from the explanation. CAIPI then uses this feedback to generate augmented data and retrain the model.

While these recent works use feedback on low-level features (input words) and individual predictions, our framework (FIND) uses feedback on the learned features with respect to the big picture of the model. This helps us avoid local decision pitfalls which usually occur in interactive machine learning \cite{wu2019local}.
Overall, what makes our contribution different from existing work is that \textit{(i)} we collect the feedback on the model, not the individual predictions, and \textit{(ii)} we target deep text classifiers which are more complex than the models used in previous work.

\section{FIND: Debugging Text Classifiers} \label{sec:met}

\begin{figure*}[t!] 
    \centering
    \includegraphics[width=0.88\linewidth]{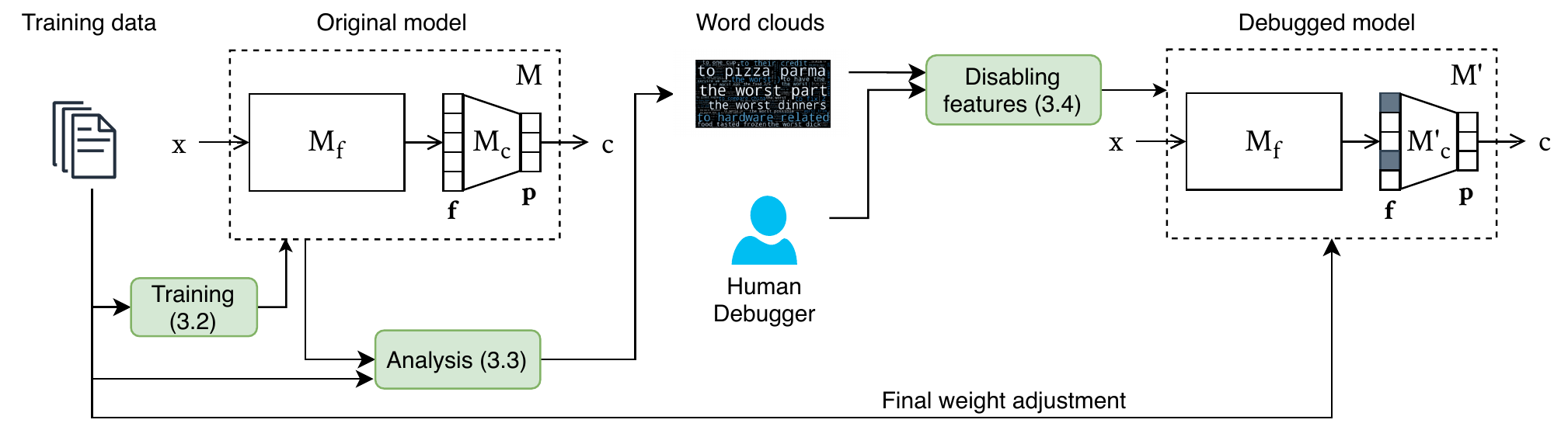}
    \caption{Overview of the proposed debugging framework, FIND. The numbers in the green boxes refer to the corresponding Sections in this paper.} \label{fig:overview}
\end{figure*}

\subsection{Motivation} \label{subsec:motivation}
Generally, deep text classifiers can be divided into two parts.
The first part performs \textit{feature extraction}, transforming an input text into a dense vector (i.e., a \textit{feature vector}) which represents the input. There are several alternatives to implement this part such as using convolutional layers, recurrent layers, and transformer layers. 
The second part performs \textit{classification} passing the feature vector through a dense layer with softmax activation to get predicted probability of the classes. 
These deep classifiers are not transparent, as humans cannot interpret the meaning of either the intermediate vectors or the model parameters used for feature extraction.
This prevents humans from applying their knowledge to modify or debug the classifiers.

In contrast, if we understand which patterns or qualities of the input are captured in each feature, we can comprehend the overall reasoning mechanism of the model as the dense layer in the classification part then becomes interpretable. In this paper, we make this possible using LRP.
By understanding the model, humans can check whether the input patterns detected by each feature are relevant for classification. 
Also, the features should be used by the subsequent dense layer to support the right classes.
If these are not the case, debugging can be done by disabling the features which may be harmful if they exist in the model.
Figure \ref{fig:overview} shows the overview of our debugging framework, FIND.

\subsection{Notation}
Let us consider a text classification task with $|\mathcal{C}|$ classes where $\mathcal{C}$ is the set of all classes and let $\mathcal{V}$ be a set of unique words in the corpus (the vocabulary).
A training dataset $\mathcal{D}=\{(x_1, y_1), \ldots, (x_N, y_N) \}$ is given, where $x_i$ is the $i$-th document containing a sequence of $L$ words, $[x_{i1}, x_{i2}, ..., x_{iL}]$, and $y_i \in \mathcal{C}$ is the class label of $x_i$.
A deep text classifier $M$ trained on  dataset $\mathcal{D}$ classifies a new input document $x$ into one of the classes (i.e., $M(x) \in \mathcal{C}$). 
In addition, $M$ can be divided into two parts -- a feature extraction part $M_f$ and a classification part $M_c$. 
Formally, $M(x) = (M_c \circ M_f)(x)$; 
$M_f(x) = \mathbf{f}$; 
$M(x) = M_c(\mathbf{f}) = \mbox{softmax}(\mathbf{Wf}+\mathbf{b}) = \mathbf{p}$ where $\mathbf{f} = [f_1, f_2, \ldots, f_d] \in \mathbb{R}^d$ is the feature vector of $x$, while $\mathbf{W} \in \mathbb{R}^{|\mathcal{C}| \times d}$ and $\mathbf{b} \in \mathbb{R}^{|\mathcal{C}|}$ are parameters of the dense layer of $M_c$. The final output is the predicted probability vector $\mathbf{p} \in [0,1]^{|\mathcal{C}|}$. 

\subsection{Understanding the Model} \label{subsec:lrp}
To understand how the model $M$ works, we analyze the patterns or characteristics of the input that activate each feature $f_i$. 
Specifically, using LRP\footnote{See Appendix \ref{app:lrp} for more details on how LRP works.}, for each $f_i$ of an example $x_j$ in the training dataset, 
we calculate a relevance vector $\mathbf{r}_{ij} \in \mathbb{R}^L$ showing the relevance scores (the contributions) of each word in $x_j$ for the value of $f_i$.  
After doing this for all $d$ features of all training examples, we can produce word clouds to help the users better understand the model $M$.

\textbf{Word clouds} -- For each feature $f_i$, we create (one or more) word clouds to visualize the patterns in the input texts which highly activate $f_i$. This can be done by analyzing $\mathbf{r}_{ij}$ for all $x_j$ in the training data and displaying, in the word clouds, words or n-grams which get high relevance scores. 
Note that different model architectures may have different ways to generate the word clouds so as to effectively reveal the behavior of the features. 

\begin{figure}[t!] 
    \centering
    \includegraphics[width=0.80\linewidth]{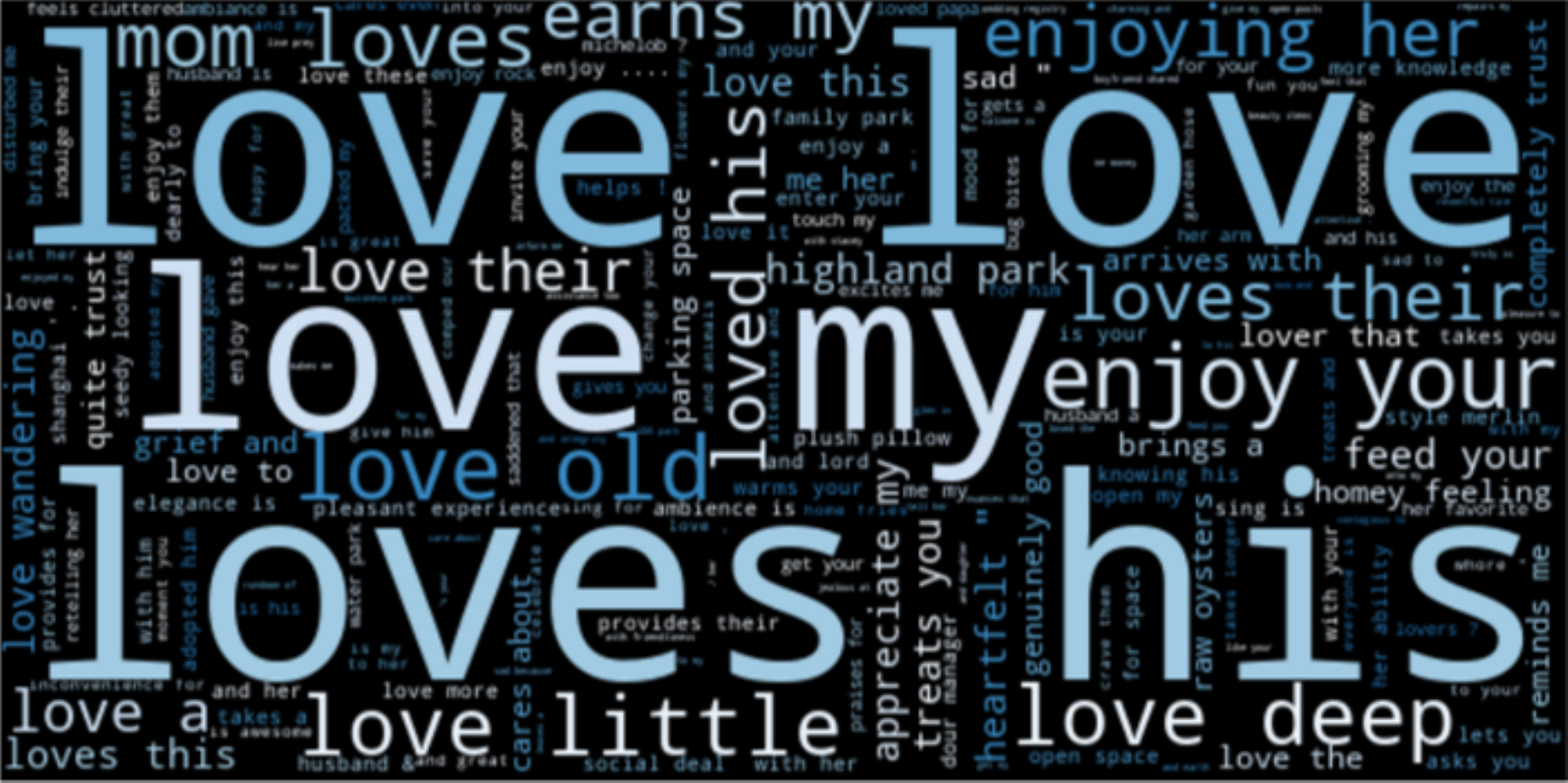}
    \caption{A word cloud (or, literally, an n-gram cloud) of a feature from a CNN.} \label{fig:wc}
\end{figure}

For CNNs, the classifiers we experiment with in this paper, each feature has one word cloud containing the n-grams, from the training examples, which were selected by the max-pooling of the CNNs.  
For instance, Figure \ref{fig:wc}, corresponding to a feature of filter size 2, shows bi-grams (e.g., ``love love'', ``love my'', ``loves his'', etc.) whose font size corresponds to the feature values of the bi-grams. 
This is similar to how previous works analyze CNN features \cite{jacovi-etal-2018-understanding,lertvittayakumjorn-toni-2019-human}, and it is equivalent to back-propagating the feature values to the input using LRP and cropping the consecutive input words with non-zero LRP scores to show in the word clouds.\footnote{We also propose how to create word clouds and perform debugging for bidirectional LSTM networks \cite{hochreiter1997long} in Appendix \ref{app:bilstm}.}

\subsection{Disabling Features} \label{subsec:disabling}
As explained earlier, we want to know whether the learned features are valid and relevant to the classification task and whether or not they get appropriate weights from the next layer. 
This is possible by letting humans consider the word cloud(s) of each feature 
and tell us which class the feature is relevant to. 
A word cloud receiving human answers that are different from the class it should support (as indicated by $\mathbf{W}$) exhibits a flaw in the model.
For example, if the word cloud in Figure \ref{fig:wc} represents the feature $f_i$ in a sentiment analysis task but the $i^{th}$ column of $\textbf{W}$ implies that
$f_i$ supports the negative sentiment class, we know the model is not correct here. 
If this word cloud appears in a product categorization task, this is also problematic because the phrases in the word cloud are not discriminative of any product category. 
Hence, we provide options for the users to disable the features which correspond to any problematic word clouds so that the features do not play a role in the classification. 
To enable this to happen, we modify $M_c$ to be $M'_c$ where 
$\mathbf{p} = M'_c(\mathbf{f}) = \mbox{softmax}((\mathbf{W}\odot\mathbf{Q})\mathbf{f}+\mathbf{b})$ and $\mathbf{Q} \in \mathbb{R}^{|\mathcal{C}|\times d}$ is a masking matrix with $\odot$ being an element-wise multiplication operator.
Initially, all elements in $\mathbf{Q}$ are ones which enable all the connections between the features and the output. To disable feature $f_i$, we set the $i^{th}$ column of $\textbf{Q}$ to be a zero vector. 
After disabling features, we then freeze the parameters of $M_f$ and fine-tune the parameters of $M'_c$ (except the masking matrix $\mathbf{Q}$) with the original training dataset $\mathcal{D}$ in the final step. 

\section{Experimental Setup} \label{sec:exp}

\begin{table}[t!]
\centering
\small
\begin{tabular}{C{0.04\textwidth} L{0.14\textwidth} C{0.03\textwidth} C{0.16\textwidth}} 
 \hline
 Exp & Dataset & $|\mathcal{C}|$ & Train / Dev / Test\\
 \hline
 \multirow{2}{*}{1} & Yelp & 2 & 500 / 100 / 38000 \\ 
 & Amazon Products & 4 & 100 / 100 / 20000 \\ \hline
 \multirow{3}{*}{2} & Biosbias & 2 & 3832 / 1277 / 1278 \\ 
 & Waseem & 2 & 10144 / 3381 / 3382 \\
 & Wikitoxic & 2 & - / - / 18965 \\
 \hline
 \multirow{5}{*}{3} & 20Newsgroups & 2 & 863 / 216 / 717 \\
 & Religion & 2 & - / - / 1819 \\
 & Amazon Clothes & 2 & 3000 / 300 / 10000 \\
 & Amazon Music & 2 & - / - / 8302\\
 & Amazon Mixed & 2 & - / - / 100000 \\
 \hline
 \end{tabular}
\caption{Datasets used in the experiments.} \label{tab:dataset}
\end{table} 

All datasets and their splits used in the experiments are listed in Table \ref{tab:dataset}. We will explain each of them in the following sections.
For each classification task, we ran and improved three models, using different random seeds, independently of one another, and the reported results are the average of the three runs.
Regarding the models, we used 1D CNNs with the same structures for all the tasks and datasets. The convolution layer had three filter sizes [2, 3, 4] with 10 filters for each size (i.e., $d = 10 \times 3 = 30$). All the activation functions were ReLU
except the softmax at the output layer. 
The input documents were padded or trimmed to have 150 words ($L=150$). We used pre-trained 300-dim GloVe vectors \cite{pennington-etal-2014-glove} as non-trainable weights in the embedding layers.
All the models were implemented using Keras and trained with Adam optimizer.
We used iNNvestigate \cite{alber2018innvestigate} to run LRP on CNN features. In particular, we used the LRP-$\epsilon$ propagation rule to stabilize the relevance scores ($\epsilon=10^{-7}$).
Finally, we used Amazon Mechanical Turk (MTurk) to collect crowdsourced responses for selecting features to disable. Each question was answered by ten workers and the answers were aggregated using majority votes or average scores depending on the question type (as explained next).  

\section{Exp 1: Feasibility Study} \label{sec:exp1}
In this feasibility study, we assessed the effectiveness of word clouds as visual explanations to reveal the behavior of CNN features.
We trained CNN models using small training datasets and evaluated the quality of CNN features based on responses from MTurk workers to the feature word clouds.
Then we disabled features based on their average quality scores. 
The assumption was: if the scores of the disabled features correlated with the drop in the model predictive performance, it meant that humans could understand and accurately assess CNN features using word clouds.
We used small training datasets so that the trained CNNs had features with different levels of quality. Some features detected useful patterns, while others overfitted the training data.

\subsection{Datasets}
We used subsets of two datasets: \textit{(1)} \textbf{Yelp} -- predicting sentiments of restaurant reviews (positive or negative) \cite{zhang2015character} and \textit{(2)} \textbf{Amazon Products} -- classifying product reviews into one of  four categories (Clothing Shoes and Jewelry, Digital Music, Office Products, or Toys and Games) \cite{he2016ups}. We sampled 500 and 100 examples to be the training data for Yelp and Amazon Products, respectively.

\subsection{Human Feedback Collection and Usage}
\begin{figure}[t!] 
    \centering
    \includegraphics[width=0.9\linewidth]{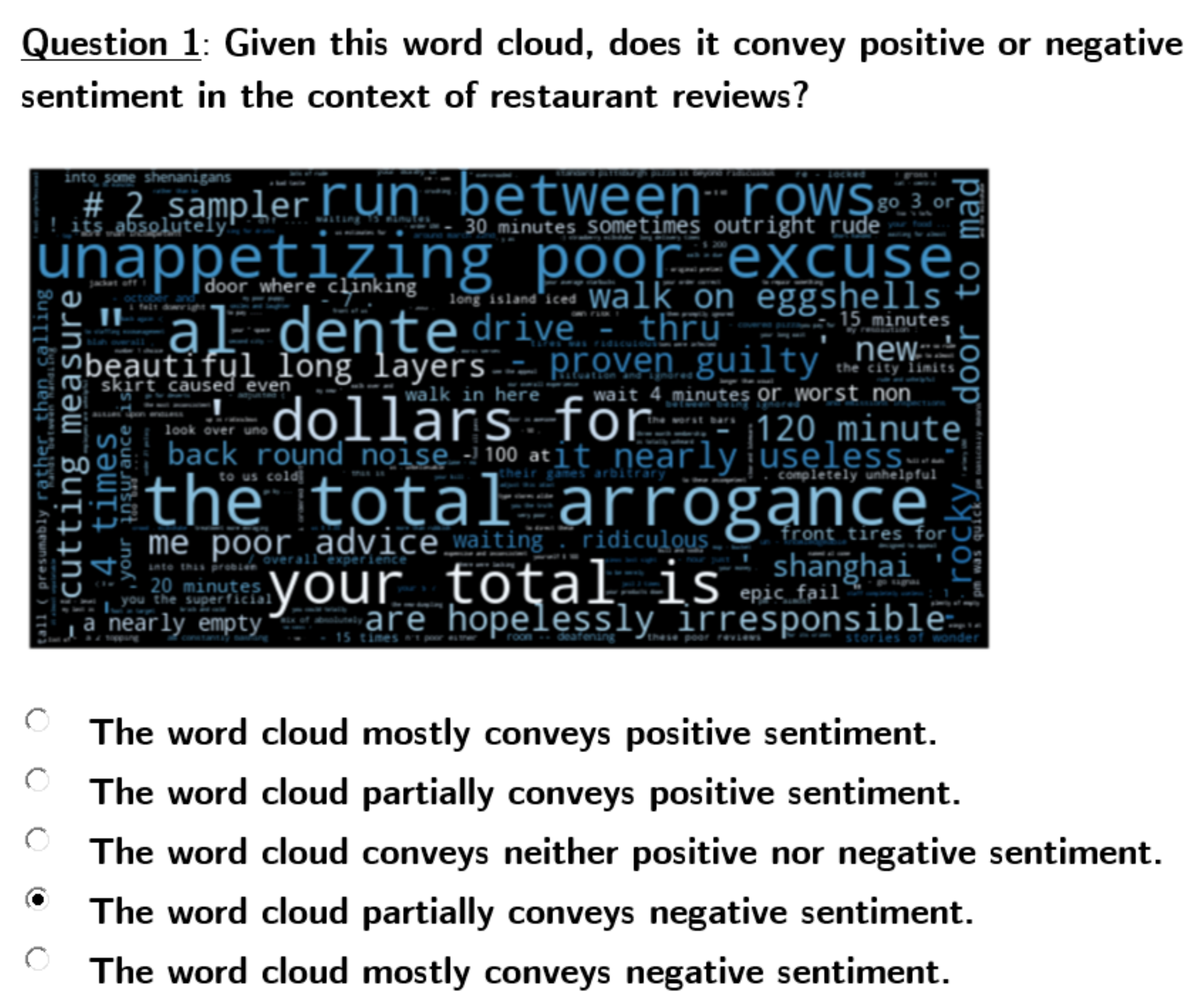}
    \caption{A user interface in Experiment 1 (Yelp).} \label{fig:ui1}
\end{figure}

We used human responses on MTurk to assign ranks to features. 
As each classifier had 30 original features ($d=30$), we divided them into three ranks (A, B, and C) each of which with 10 features. We expected that features in rank A are most relevant and useful for the prediction task, and features in rank C  least relevant, potentially undermining the performance of the model. 
To make the annotation more accessible to lay users, we designed the questions to ask whether a given word cloud is (mostly or partially) relevant to one of the classes or not, as shown in Figure \ref{fig:ui1}. 
If the answer matches how the model really uses this feature (as indicated by $\mathbf{W}$),
the feature gets a positive score from this human response. 
For example, if the CNN feature of the word cloud in Figure \ref{fig:ui1} is used by the model for the negative sentiment class, the scores of the five options in the figure are -2, -1, 0, 1, 2, respectively. We collected ten responses for each question and used the average score to sort the features descendingly. After sorting, the 1\textsuperscript{st}-10\textsuperscript{th} features, 11\textsuperscript{th}-20\textsuperscript{th} features, and 21\textsuperscript{st}-30\textsuperscript{th} features are considered as rank A, B, and C, respectively.\footnote{The questions and scoring criteria for the Amazon Products dataset, which is a multiclass classification task, are slightly different. See Appendix \ref{app:multiclass} for details.}  To show the effects of feature disabling, we compared the original model $M$ with the modified model $M'$ with features in rank X disabled where X $\in \{\mbox{A, B, C, A and B, A and C, B and C}\}$. 

\subsection{Results and Discussions}
\begin{figure}[t!] 
    \centering
    \includegraphics[width=0.98\linewidth]{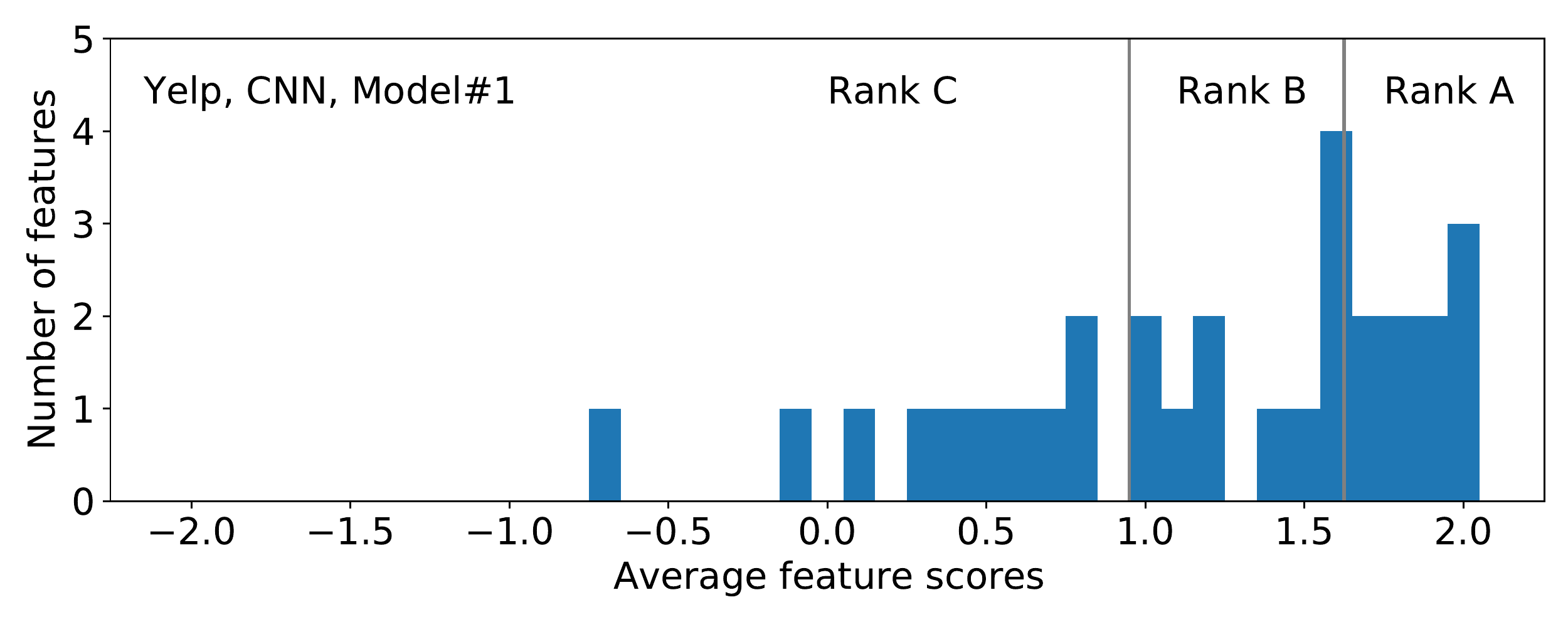}
    \caption{The distribution of average feature scores in a CNN model trained on the Yelp dataset.} \label{fig:res1}
\end{figure}

\begin{figure}[t!] 
    \centering
    \includegraphics[width=0.80\linewidth]{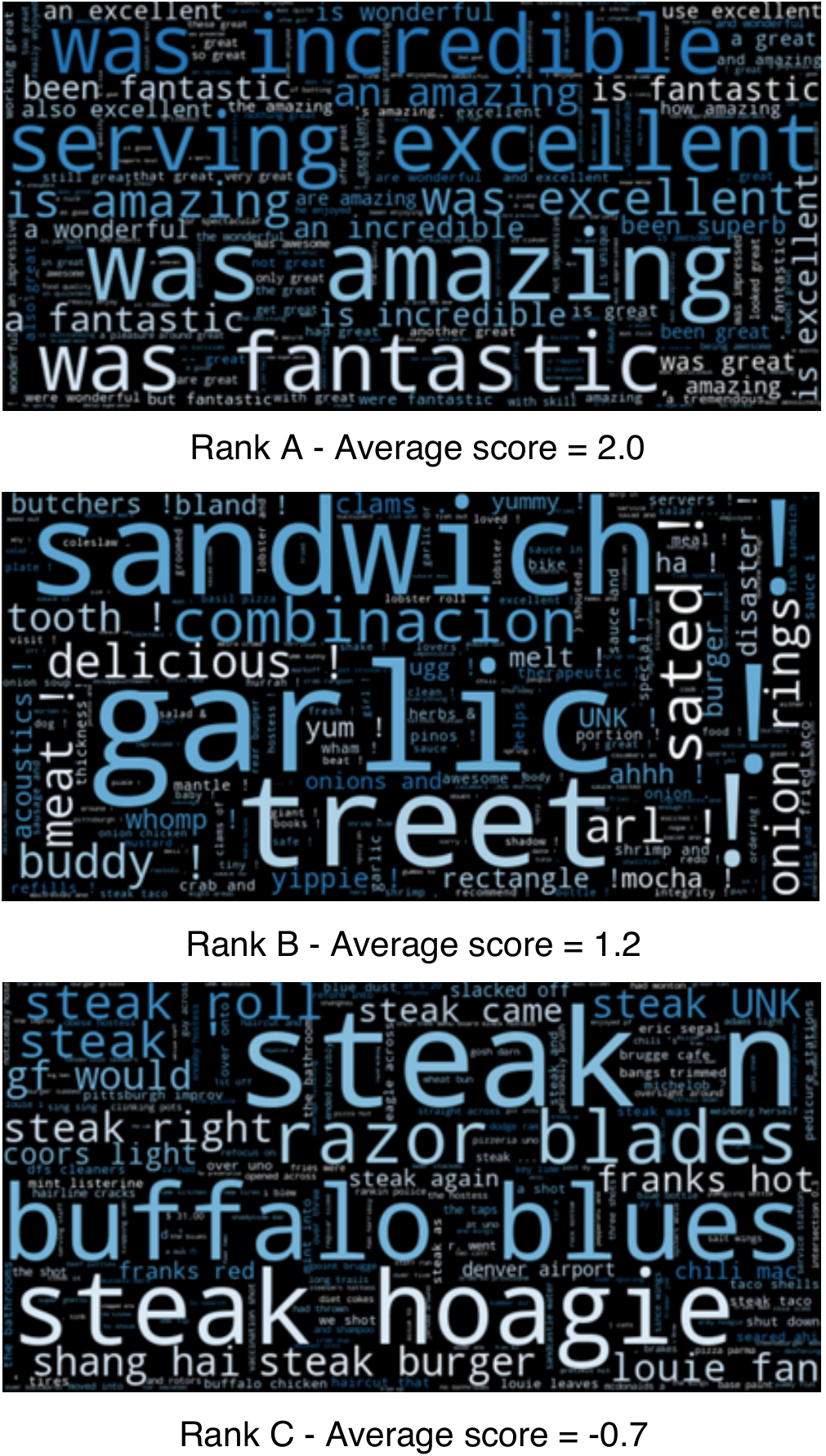}
    \caption{Examples of word clouds of CNN features in ranks A, B, and C (Experiment 1, Yelp -- sentiment).} \label{fig:wcres1}
\end{figure}

\begin{figure}[t!] 
    \centering
    \includegraphics[width=0.97\linewidth]{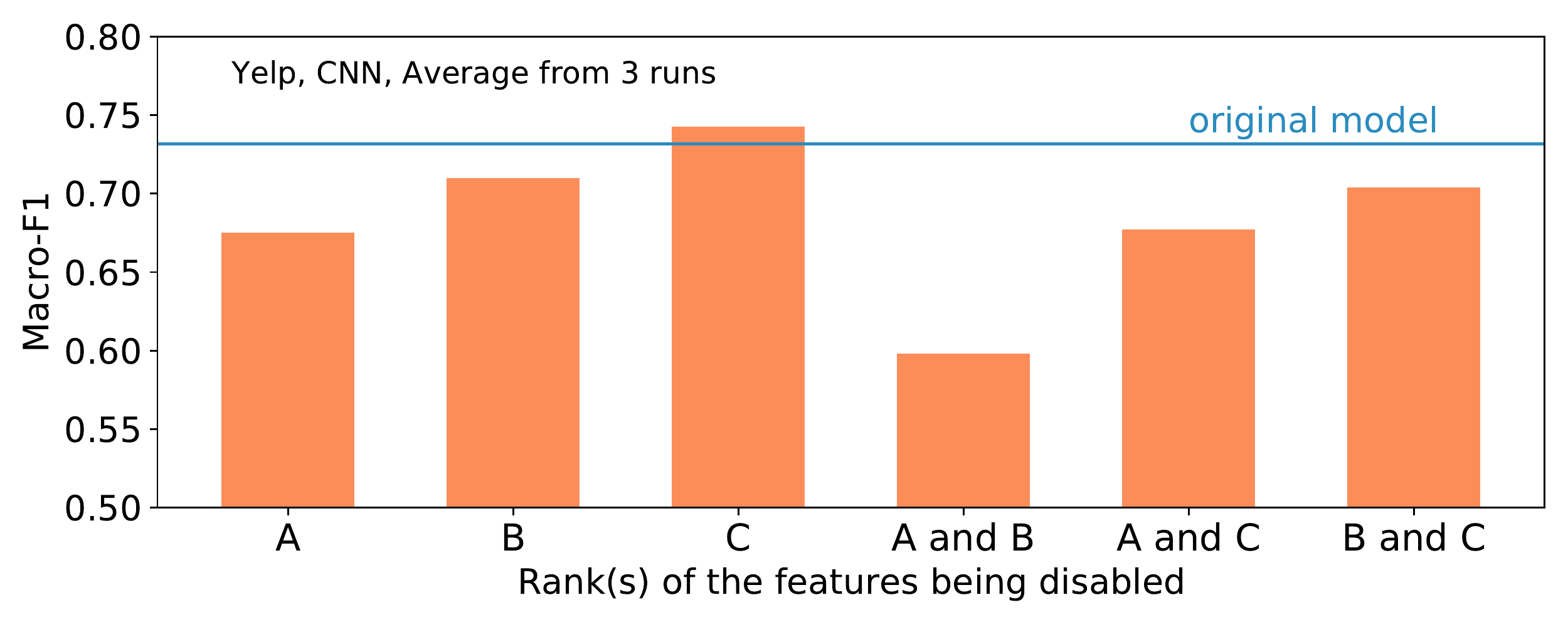}
    \includegraphics[width=0.97\linewidth]{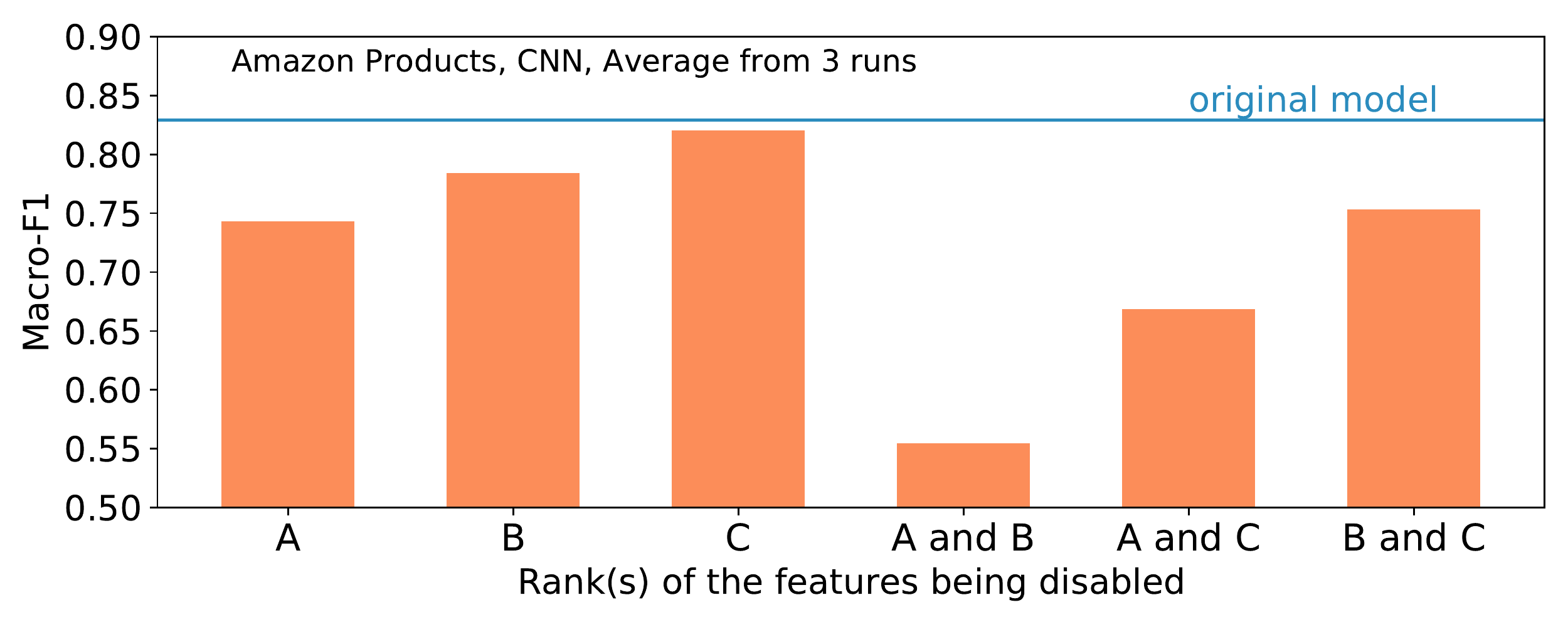}
    \caption{The average macro F1, from the three runs, of all the CNN models 
    for the Yelp dataset (top) and the Amazon Products dataset (bottom).} \label{fig:res2}
\end{figure}

Figure \ref{fig:res1} shows the distribution of average feature scores from one of the three CNN instances for the Yelp dataset. 
Examples of the word clouds from each rank are displayed in Figure \ref{fig:wcres1}. We can clearly see dissimilar qualities of the three features.
Some participants answered that the rank B feature in Figure \ref{fig:wcres1} was relevant to the positive class (probably due to the word `delicious'), and the weights of this feature in $\mathbf{W}$ agreed (Positive:Negative = 0.137:-0.135).
Interestingly, the rank C feature in Figure \ref{fig:wcres1} got a negative score because some participants believed that this word cloud was relevant to the positive class, but actually the model used this feature as evidence for the negative class (Positive:Negative = 0.209:0.385). 

Considering all the three runs, Figure \ref{fig:res2} (top) shows the average macro F1 score of the original model (the blue line) and of each modified model. 
The order of the performance drops is AB $>$ A $>$ AC $>$ BC $>$ B $>$ Original $>$ C. 
This makes sense because disabling important features (rank A and/or B) caused larger performance drops, and the overall results are consistent with the average feature scores given by the participants (as in Figure \ref{fig:res1}).
It confirms that using word clouds is an effective way to assess CNN features.
Also, it is worth noting that the macro F1 of the model slightly increased when we disabled the low-quality features (rank C). 
This shows that humans can improve the model by disabling irrelevant features. 

The CNNs for the Amazon Products dataset also behaved in a similar way (Figure \ref{fig:res2} -- bottom), except that disabling rank C features slightly undermined, not increased, performance. This implies that even the rank C features contain a certain amount of useful knowledge for this classifier.\footnote{We also conducted the same experiments here with bidirectional LSTM networks (BiLSTMs) which required a different way to generate the word clouds 
(see Appendix \ref{app:bilstm}).
The results on BiLSTMs, however, are not as promising as on CNNs. This might be because the way we created word clouds for each BiLSTM feature was not an accurate way to reveal its behavior.
Unlike for CNNs, understanding recurrent neural network features for text classification is still an open problem.}

\section{Exp 2: Training Data with Biases} \label{sec:exp2}
Given a biased training dataset, a text classifier may absorb the biases and produce biased predictions against some sub-populations. 
We hypothesize that if the biases are captured by some of the learned features, we can apply FIND to disable such features and reduce the model biases.

\subsection{Datasets and Metrics}
We focus on reducing gender bias of CNN models trained on two datasets -- \textbf{Biosbias} \cite{dearteaga2019biosbias} and \textbf{Waseem} \cite{waseem-hovy-2016-hateful}. 
For Biosbias, the task is predicting the occupation of a given bio paragraph, i.e., whether the person is `a surgeon' (class 0) or `a nurse' (class 1). 
Due to the gender imbalance in each occupation, a classifier usually exploits gender information when making predictions. 
As a result, bios of female surgeons and male nurses are often misclassified. 
For Waseem, the task is abusive language detection -- assessing if a given text is abusive (class 1) or not abusive (class 0). 
Previous work found that this dataset contains a strong negative bias against females \cite{park-etal-2018-reducing}. 
In other words, texts related to females are usually classified as abusive although the texts themselves are not abusive at all.
Also, we tested the models, trained on the Waseem dataset, using another abusive language detection dataset, \textbf{Wikitoxic} \cite{wikitoxic}, to assess generalizability of the models.
To quantify gender biases, we adopted two metrics -- false positive equality difference (FPED) and false negative equality difference (FNED) \cite{dixon2018measuring}. 
The lower these metrics are, the less biases the model has.

\subsection{Human Feedback Collection and Usage}
Unlike the interface in Figure \ref{fig:ui1}, for each word cloud, we asked the participants to select the relevant class from three options (Biosbias: surgeon, nurse, it could be either / Waseem: abusive, non-abusive, it could be either). The feature will be disabled if the majority vote does not select the class suggested by the weight matrix $\mathbf{W}$. 
To ensure that the participants do not use their biases while answering our questions, we firmly mentioned in the instructions that gender-related terms should not be used as an indicator for one or the other class.

\subsection{Results and Discussions}
\begin{figure}[t!] 
    \centering
    \includegraphics[width=0.97\linewidth]{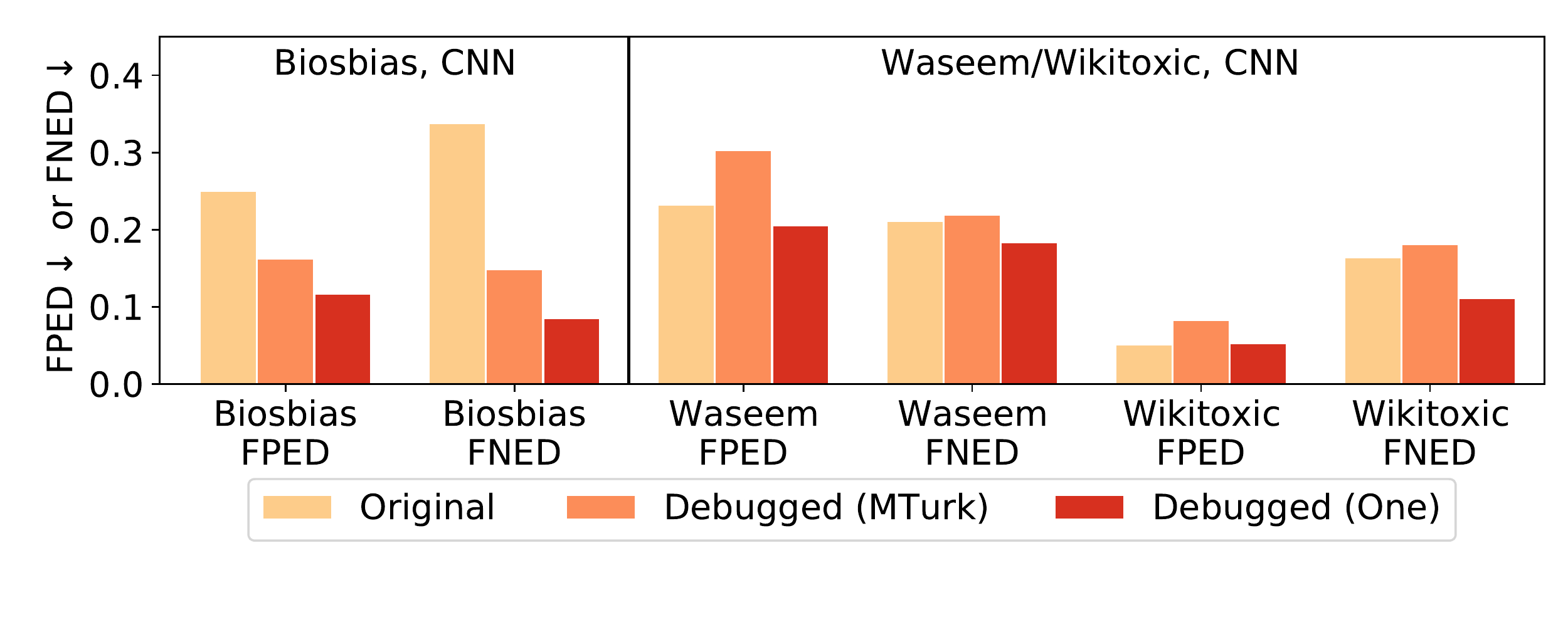}
    \caption{The average FPED and FNED of the CNN models 
    in Experiment 2 (the lower, the better).} \label{fig:waseem}
\end{figure}

The results of this experiment are displayed in Figure \ref{fig:waseem}.
For Biosbias, on average, the participants' responses suggested us to disable 11.33 out of 30 CNN features. By doing so, the FPED of the models decreased from 0.250 to 0.163, and the FNED decreased from 0.338 to 0.149. 
After investigating the word clouds of the CNN features, we found that some of them detected patterns containing both gender-related terms and occupation-related terms such as ``his surgical expertise'' and ``she supervises nursing students''. 
Most of the MTurk participants answered that these word clouds were relevant to the occupations, and thus the corresponding features were not disabled.
However, we believe that these features might contain gender biases. 
So, we asked one annotator to consider all the word clouds again and disable every feature for which the prominent n-gram patterns contained any gender-related terms, no matter whether the patterns detect occupation-related terms.
With this new disabling policy, 12 out of 30 features were disabled on average, and the model biases further decreased, as shown in Figure \ref{fig:waseem} (Debugged (One)).
The side-effect of disabling 33\% of all the features here was only a slight drop in the macro F1 from 0.950 to 0.933. Hence, our framework was successful in reducing gender biases without severe negative effects in classification performance.

Concerning the abusive language detection task, on average, the MTurk participants' responses suggested us to disable 12 out of 30 CNN features.
Unlike Biosbias, disabling features based on MTurk responses unexpectedly increased the gender bias for both Waseem and Wikitoxic datasets. 
However, we found one similar finding to Biosbias, that many of the CNN features captured n-grams which were both abusive and related to a gender such as `these girls are terrible' and `of raping slave girls', and these features were not yet disabled. 
So, we asked one annotator to disable the features using the new ``brutal'' policy -- disabling all which involved gender words even though some of them also detected abusive words. 
By disabling 18 out of 30 features on average, the gender biases were reduced for both datasets (except FPED on Wikitoxic which stayed close to the original value).
Another consequence was that we sacrificed 4\% and 1\% macro F1 on the Waseem and Wikitoxic datasets, respectively. 
This finding is consistent with \cite{park-etal-2018-reducing} that reducing the bias and maintaining the classification performance at the same time is very challenging.

\section{Exp 3: Dataset Shift} \label{sec:exp3}
Dataset shift is a problem where the joint distribution of inputs and outputs differs between training and test stage \cite{datasetshift}.
Many classifiers perform poorly under dataset shift because some of the learned features are inapplicable (or sometimes even harmful) to  classify test documents.
We hypothesize that FIND is useful for investigating the learned features and disabling the overfitting ones to increase the generalizability of the model.

\subsection{Datasets}
We considered two tasks in this experiment. The first task aims to classify ``Christianity'' vs ``Atheism'' documents from the \textbf{20 Newsgroups} dataset\footnote{\url{http://qwone.com/~jason/20Newsgroups/}}. 
This dataset is special because it contains a lot of artifacts -- tokens (e.g., person names, punctuation marks) which are not relevant, but strongly co-occur with one of the classes. 
For evaluation, we used the \textbf{Religion} dataset by \citet{ribeiro2016lime}, containing ``Christianity'' and ``Atheism'' web pages, as a target dataset.
The second task is sentiment analysis. We used, as a training dataset, \textbf{Amazon Clothes}, with reviews of clothing, shoes, and jewelry products \cite{he2016ups}, and as test sets  three out-of-distribution datasets -- \textbf{Amazon Music} \cite{he2016ups}, \textbf{Amazon Mixed} \cite{zhang2015character}, and the \textbf{Yelp} dataset (which was used in Experiment 1). 
Amazon Music contains only reviews from the ``Digital Music'' product category which was found to have an extreme distribution shift from the clothes category \cite{hendrycks2020pretrained}.
Amazon Mixed compiles the reviews from various kinds of products, while Yelp focuses on restaurant reviews.

\subsection{Human Feedback Collection and Usage}
We collected responses from MTurk workers using the same user interfaces as in Experiment 2. Simply put, we asked the workers to select a class which was relevant to a given word cloud and checked if the majority vote agreed with the weights in $\mathbf{W}$.

\subsection{Results and Discussions}
\begin{figure}[t!] 
    \centering
    \includegraphics[width=0.97\linewidth]{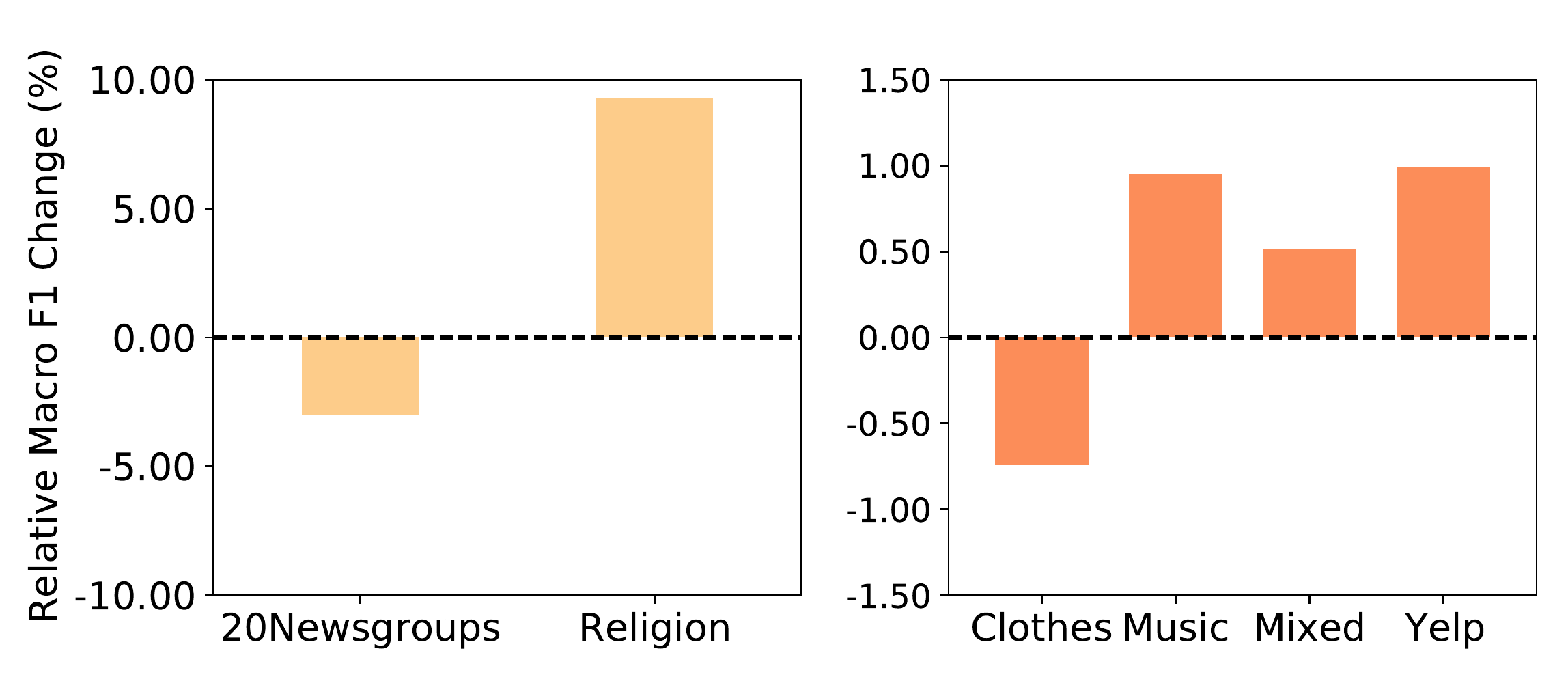}
    \caption{The relative Macro F1 changes (in \%) of the CNN models for both tasks in Experiment 3.} \label{fig:clothes}
\end{figure}

For the first task, on average, 14.33 out of 30 features were disabled and the macro F1 scores of the 20Newsgroups before and after debugging are 0.853 and 0.828, respectively. 
The same metrics of the Religion dataset are 0.731 and 0.799. 
This shows that disabling irrelevant features mildly undermined the predictive performance on the in-distribution dataset, but clearly enhanced the performance on the out-of-distribution dataset (see Figure \ref{fig:clothes}, left). 
This is especially evident for the Atheism class for which the F1 score increased around 15\% absolute. 
We noticed from the word clouds that 
many prominent words for the Atheism class learned by the models are person names (e.g., Keith, Gregg, Schneider) and these are not applicable to the Religion dataset. Forcing the models to use only relevant features (detecting terms like `atheists' and `science'), therefore, increased the macro F1 on the Religion dataset.   

Unlike 20Newsgroups, Amazon Clothes does not seem to have obvious artifacts. Still, the responses from crowd workers suggested that we disable 6 features. 
The disabled features were correlated to, but not the reason for, the associated class. 
For instance, one of the disabled features was highly activated by the pattern ``my .... year old'' which often appeared in positive reviews such as ``my 3 year old son loves this.''. 
However, these correlated features are not very useful for the three out-of-distribution datasets (Music, Mixed, and Yelp). 
Disabling them made the model focus more on the right evidence and increased the average macro F1 for the three datasets, as shown in Figure \ref{fig:clothes} (right). 
Nonetheless, the performance improvement here was not as apparent as in the previous task because, even without feature disabling, the majority of the features are relevant to the task and can lead the model to the correct predictions in most cases.\footnote{See Appendix \ref{app:results} for the full results from all  experiments.}

\section{Discussion and Conclusions} \label{sec:con}
We proposed FIND, a framework which enables humans to debug deep text classifiers by disabling irrelevant or harmful features. Using the proposed framework on CNN text classifiers, we found that 
\textit{(i)} word clouds generated by running LRP on the training data accurately revealed the behaviors of CNN features,    
\textit{(ii)} some of the learned features might be more useful to the task than the others
and \textit{(iii)} disabling the irrelevant or harmful features could improve the model predictive performance and reduce unintended biases in the model.

\subsection{Generalization to Other Models}
In order to generalize the framework beyond CNNs, there are two questions to consider. 
First, what is an effective way to understand each feature? We exemplified this with two word clouds representing each BiLSTM feature in Appendix \ref{app:bilstm}, and we plan to experiment with advanced visualizations such as LSTMVis \cite{strobelt2018lstmvis} in the future. 
Second, can we make the model features more interpretable? 
For example, using ReLU as activation functions in LSTM cells (instead of tanh) renders the features non-negative. So, they can be summarized using one word cloud which is more practical for debugging.

In general, the principle of FIND is understanding the features and then disabling the irrelevant ones. The process makes \textit{visualizations} and \textit{interpretability} more actionable. 
Over the past few years, we have seen rapid growth of scientific research in both topics (visualizations and interpretability) aiming to understand many emerging advanced models including the popular transformer-based models \cite{jo-myaeng-2020-roles,voita-etal-2019-analyzing,hoover-etal-2020-exbert}. We believe that our work will inspire other researchers to foster advances in both topics towards the more tangible goal of model debugging.

\subsection{Generalization to Other Tasks}
FIND is suitable for any text classification tasks where a model might learn irrelevant or harmful features during training. It is also convenient to use since only the trained model and the training data are required as input. 
Moreover, it can address many problems simultaneously such as removing religious and racial bias together with gender bias even if we might not be aware of such problems before using FIND. 
In general cases, FIND is at least useful for model verification.

For future work, it would be interesting to extend FIND to other NLP tasks, e.g., question answering and natural language inference. This will require some modifications to understand how the features capture relationships between two input texts.

\subsection{Limitations}
Nevertheless, FIND has some limitations. 
First, the word clouds may reveal sensitive contents in the training data to human debuggers. 
Second, the more hidden features the model has, the more human effort FIND needs for debugging. For instance, BERT-base \cite{devlin-etal-2019-bert} has 768 features (before the final dense layer) which require lots of human effort to perform investigation. In this case, it would be more efficient to use FIND to disable attention heads rather than individual features \cite{voita-etal-2019-analyzing}.
Third, it is possible that one feature detects several patterns \cite{jacovi-etal-2018-understanding} and it will be difficult to disable the feature if some of the detected patterns are useful while the others are harmful. Hence, FIND would be more effective when used together with disentangled text representations \cite{cheng-etal-2020-improving}. 

\section*{Acknowledgments}
We would like to thank Nontawat Charoenphakdee and anonymous reviewers for helpful comments. 
Also, the first author wishes to thank the support from Anandamahidol Foundation, Thailand.

% The acknowledgments should go immediately before the references. Do not number the acknowledgments section.
% Do not include this section when submitting your paper for review.

\bibliography{anthology,emnlp2020}
\bibliographystyle{acl_natbib}

\appendix

\section{Layer-wise Relevance Propagation} \label{app:lrp}
Layer-wise Relevance Propagation (LRP) is a technique for explaining predictions of neural networks in terms of importance scores of input features \cite{bach2015lrp}. Originally, it was devised to explain predictions of image classifiers by creating a heatmap on the input image highlighting pixels that are important for the classification. Then \citet{arras-etal-2016-explaining} and \citet{arras-etal-2017-explaining} extended LRP to work on CNNs and RNNs for text classification, respectively.

Consider a neuron $k$ whose value is computed using $n$ neurons in the previous layer,
\[x_k = g(\sum_{j=1}^nx_jw_{jk}+b_k)\]
where $x_k$ is the value of the neuron $k$, $g$ is a non-linear activation function, $w_{jk}$ and $b_k$ are weights and bias in the network, respectively. We can see that the contribution of a single node $j$ to the value of the node $k$ is
\[z_{jk} = x_jw_{jk}+\frac{b_k}{n}\]
assuming that the bias term $b_k$ is distributed equally to the $n$ neurons. LRP works by propagating the activation of a neuron of interest  back through the previous layers in the network proportionally. We call the value each neuron receives a relevance score ($R$) of the neuron. To back propagate, if the relevance score of the neuron $k$ is $R_k$, the relevance score that the neuron $j$ receives from the neuron $k$ is
\[R_{j\leftarrow k}=\frac{z_{jk}}{\sum_{j'=1}^{n}z_{j'k}}R_k\]
To make the relevance propagation more stable, we add a small positive number $\epsilon$ (as a stabilizer) to the denominator of the propagation rule:
\[R_{j\leftarrow k}=\frac{z_{jk}}{\epsilon+\sum_{j'=1}^{n}z_{j'k}}R_k\]
We used this propagation rule, so called LRP-$\epsilon$, in the experiments of this paper. For more details about LRP propagation rules, please see \citet{montavon2019layer}.

To explain a prediction of a CNN text classifier, we propagate an activation value of the output node back to the word embedding matrix. After that, the relevance score of an input word equals the sum of relevance scores each dimension of its word vector receives. However, in this paper, we want to analyze the hidden features rather than the output, so we start back propagating from the hidden features instead to capture patterns of input words which highly activate the features.

\section{Multiclass Classification} \label{app:multiclass}

\begin{figure}[t!] 
    \centering
    \includegraphics[width=\linewidth]{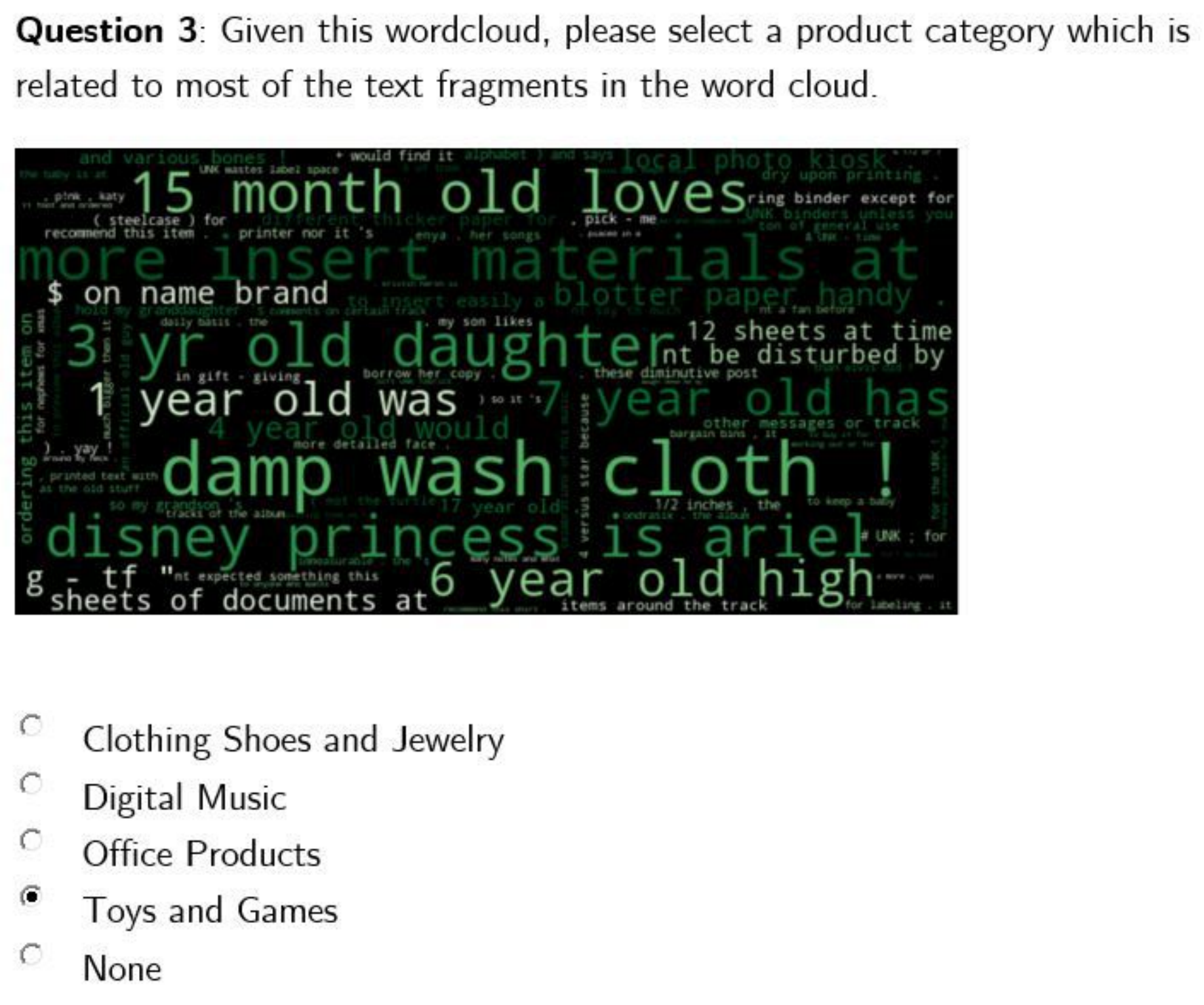}
    \caption{A user interface in Experiment 1 (Amazon Products).} \label{fig:app-ui1b}
\end{figure}

\begin{figure}[t!] 
    \centering
    \includegraphics[width=\linewidth]{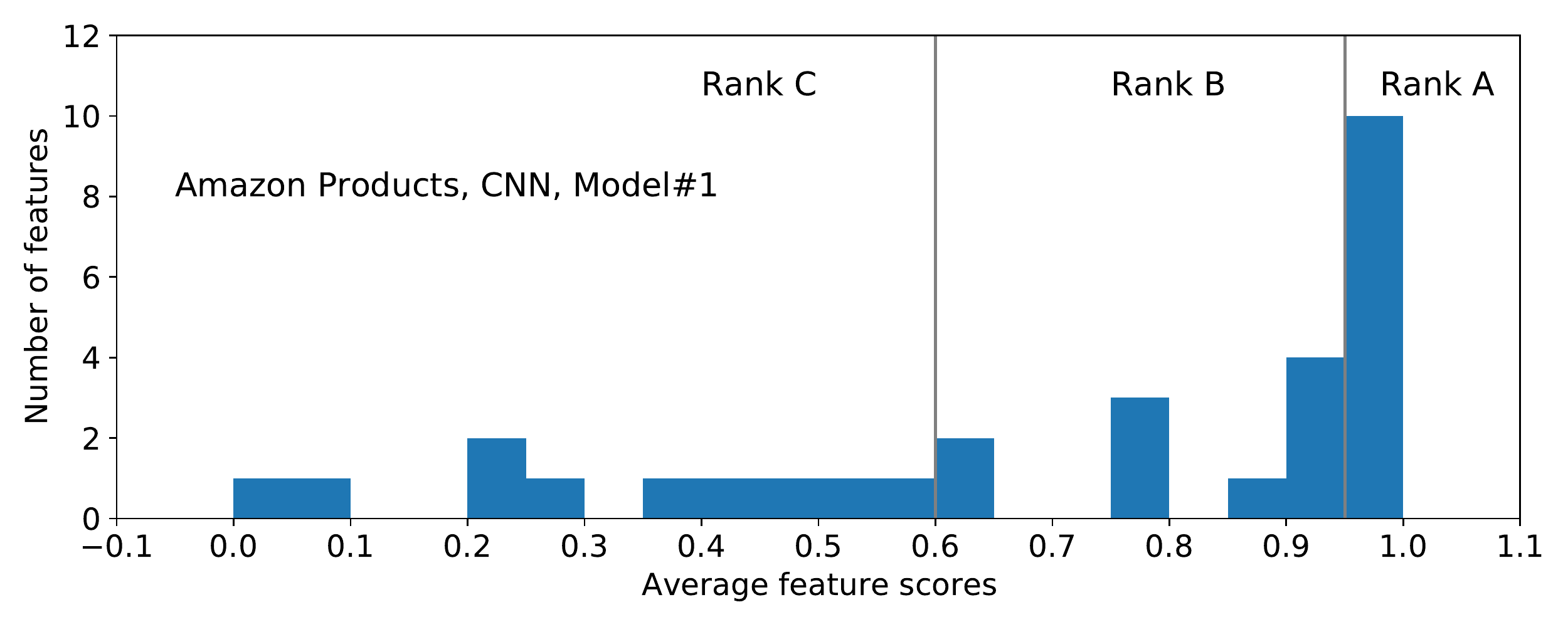}
    \caption{The distribution of average feature scores in a CNN model trained on the Amazon Products dataset.} \label{fig:app-dist1b}
\end{figure}

As shown in Figure \ref{fig:app-ui1b}, we used a slightly different user interface in Experiment 1 for the Amazon Products dataset which is a multiclass classification task. In this setting, we did not provide the options for mostly and partly  relevant; otherwise, there would have been nine options per question which are too many for the participants to answer accurately. 
With the user interface in Figure \ref{fig:app-ui1b}, we gave a score to the feature $f_i$ based on the participant answer. To explain, we re-scaled values in 
the $i^{th}$ column of $\mathbf{W}$
to be in the range [0,1] using min-max normalization and gave the normalized value of the chosen class as a score to the feature $f_i$. If the participant selects None, this feature gets a zero score.
The distribution of the average feature scores for this task (one CNN) is displayed in Figure \ref{fig:app-dist1b}.

\section{Bidirectional LSTM networks} \label{app:bilstm}
\begin{figure}[t!] 
    \centering
    \includegraphics[width=\linewidth]{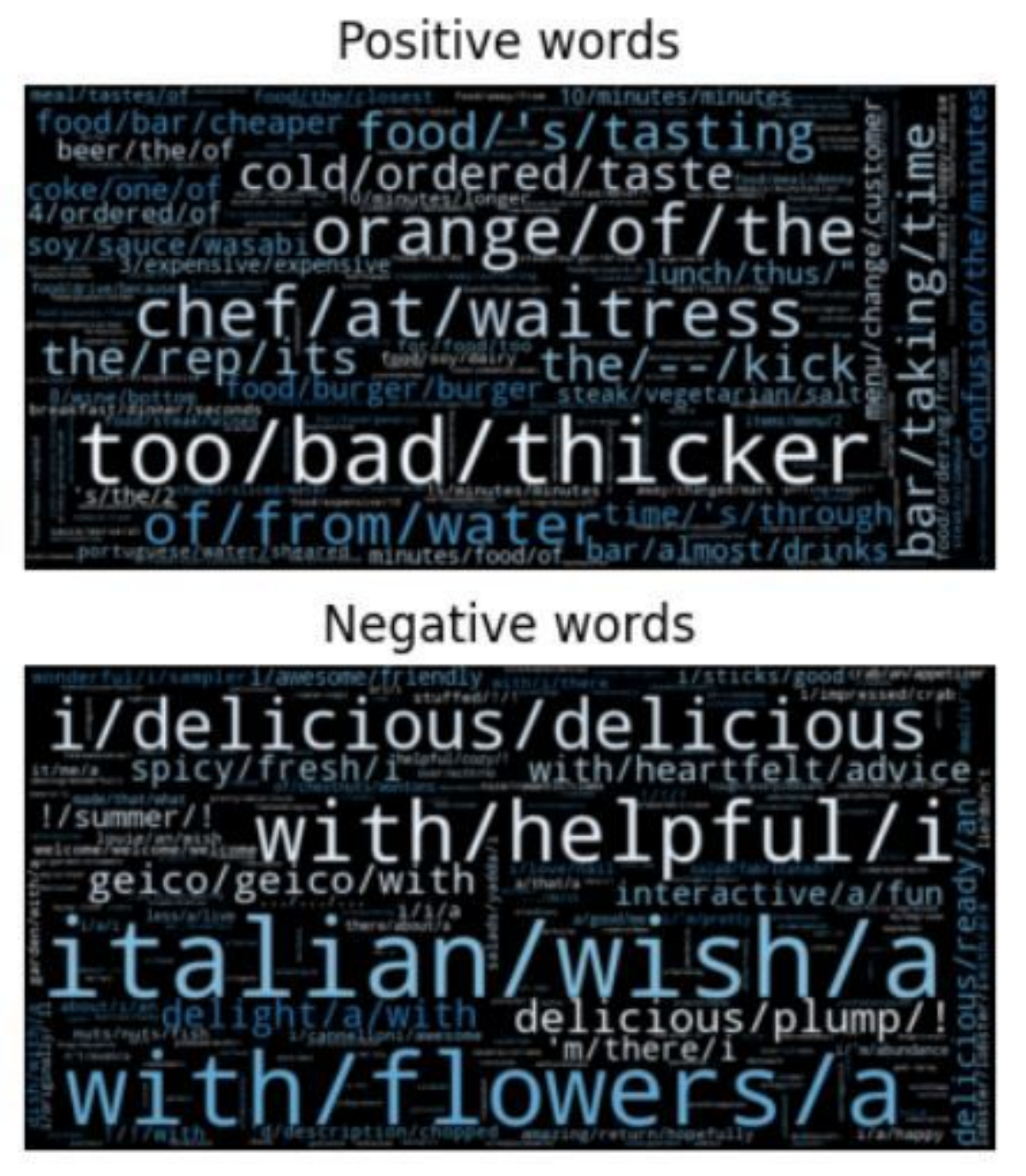}
    \caption{A pair of word clouds which represent one BiLSTM feature.} \label{fig:app-bilstmwc}
\end{figure}

To understand BiLSTM features, we created two word clouds for each feature. The first word cloud contains top three words which gain the highest positive relevance scores from each training example, while the second word cloud does the same but for the top three words which gain the lowest negative relevance scores (see Figure \ref{fig:app-bilstmwc}). 

Furthermore, we also conducted Experiment 1 for BiLSTMs. 
Each direction of the recurrent layer had 15 hidden units and the feature vector was obtained by taking element-wise max of all the hidden states (i.e., $d = 15 \times 2 = 30$).
We adapted the code of \cite{arras-etal-2017-explaining} to run LRP on BiLSTMs. 
Regarding human feedback collection, we collected feedback from Amazon Mechanical Turk workers by splitting the pair of word clouds into two and asking the question about the relevant class independently of each other. The answer of the positive relevance word cloud should be consistent with the weight matrix $\mathbf{W}$, while the answer of the negative relevance word cloud should be the opposite of the weight matrix $\mathbf{W}$. 
The score of a BiLSTM feature is the sum of its scores from the positive word cloud and the negative word cloud.  

The results of the extra BiLSTM experiments are shown in Table \ref{tab:1c} and \ref{tab:1d}. Table \ref{tab:1c} shows unexpected results after disabling features. For instance, disabling rank B features caused a larger performance drop than removing rank A features. This suggests that how we created word clouds for each BiLSTM feature (i.e., displaying top three words with the highest positive and lowest negative relevance) might not be an accurate way to explain the feature. Nevertheless, another observation from Table \ref{tab:1c} is that even when we disabled two-third of the BiLSTM features, the maximum macro F1 drop was less than 5\%. This suggests that there is a lot of redundant information in the features of the BiLSTMs.

\section{Metrics for Biases}
In this paper, we used two metrics to quantify biases in the models -- False positive equality difference (FPED) and False negative equality difference (FNED) -- with the following definitions \cite{dixon2018measuring}.
$$FPED = \sum_{t \in T}|FPR - FPR_t|$$
$$FNED = \sum_{t \in T}|FNR - FNR_t|$$
where $T$ is a set of all sub-populations we consider (i.e., $T=\{\mbox{male}, \mbox{female}\}$). FPR and FNR stand for false positive rate and false negative rate, respectively. The subscript $t$ means that we calculate the metrics using data examples mentioning the sub-population $t$ only. 
We used the following keywords to identify examples which are related to or mentioning the sub-populations.

\noindent\textbf{Male gender terms:}

``male", ``males", ``boy", ``boys", ``man", ``men", 
``gentleman", ``gentlemen", ``he", ``him", ``his", ``himself", ``brother", ``son", ``husband", ``boyfriend", ``father", ``uncle", ``dad"

\noindent\textbf{Female gender terms:}

``female", ``females", ``girl", ``girls", ``woman", ``women", 
``lady", ``ladies", ``she", ``her", ``herself", ``sister", ``daughter",
``wife", ``girlfriend", ``mother", ``aunt", ``mom"

\section{Additional Details for Reproducibility}
\subsection{Data Sources and Pre-processing}
\begin{itemize}
    \item \textbf{Yelp} and \textbf{Amazon Mixed}: We sampled examples from the datasets provided by \citet{zhang2015character} here\footnote{\url{https://github.com/zhangxiangxiao/Crepe}}.
    \item \textbf{Amazon Products}, \textbf{Amazon Clothes}, \textbf{Amazon Music}: We sampled examples from the datasets provided by \citet{he2016ups} here\footnote{\url{http://jmcauley.ucsd.edu/data/amazon/}}.
    \item \textbf{Biosbias}: We created the dataset using the code provided by \citet{dearteaga2019biosbias} here\footnote{\url{https://github.com/Microsoft/biosbias}}. All the bios are from Common Crawl August 2018 Index.
    \item \textbf{Waseem}: The authors of \cite{waseem-hovy-2016-hateful} kindly provided the dataset to us by email. We considered ``racism'' and ``sexism'' examples as ``Abusive'' and ``neither'' examples as ``Non-abusive''.
    \item \textbf{Wikitoxic}: The dataset can be downloaded here\footnote{\url{https://figshare.com/articles/Wikipedia_Talk_Labels_Toxicity/4563973}}. We used only examples which were given the same label by all the annotators.
    \item \textbf{20Newsgroups}: We downloaded the standard splits of the dataset using scikit-learn\footnote{\url{https://scikit-learn.org/}}. The header and the footer of each text were removed.
    \item \textbf{Religion}: We used the dataset provided by \citet{ribeiro2016lime} here\footnote{\url{https://github.com/marcotcr/lime-experiments}}.
\end{itemize}

\subsection{Number of Model Parameters}
\textbf{Convolutional Neural Networks}
\begin{itemize}
    \setlength\itemsep{0em}
    \item Fixed word embeddings: 120,000,600
    \item Convolutional layers: 27,030
    \item Final (masked) dense layer:
    \begin{itemize}
        \item Binary classification: 62 (+60)
        \item 4-class classification: 124 (+120)
    \end{itemize}
\end{itemize}

\noindent \textbf{Bidirectional LSTM networks}
\begin{itemize}
    \setlength\itemsep{0em}
    \item Fixed word embeddings: 120,000,600
    \item Bidirectional LSTM layers: 37,920
    \item Final (masked) dense layer:
    \begin{itemize}
        \item Binary classification: 62 (+60)
        \item 4-class classification: 124 (+120)
    \end{itemize}
\end{itemize}

\subsection{Computing Infrastructure Used}
\begin{itemize}
    \setlength\itemsep{0em}
    \item CPU: Intel Core i9-9900X (3.5GHz)
    \item GPU: 11GB NVIDIA GeForce RTX 2080 Ti
    \item RAM: 32GB Corsair Vengeance DDR4
\end{itemize}

\section{Full Experimental Results} \label{app:results}
Tables~\ref{tab:1a}-\ref{tab:3b} in this section report the full results of all the experiments and datasets. All the results shown are averaged from three runs. Boldface numbers are the best scores in the columns. They are further underlined if they are significantly better than the scores of all the other models.
We conducted the statistical significance analysis using approximate randomization test with 1,000 iterations and a significance level $\alpha$ of 0.05 \cite{noreen1989computer,graham-etal-2014-randomized}.

\begin{table*}[t!] 
    \centering
    \small
    \begin{tabular}{|l|c|c|c|c|}
    \hline
    \multirow{2}{*}{\makecell[c]{Model: CNNs}} & \multicolumn{4}{c|}{\textbf{Test dataset: Yelp}} \\ \cline{2-5}
    & Negative F1 & Positive F1 & Accuracy & Macro F1 \\ \hline
    Original & \textbf{\underline{0.758 $\pm$ 0.04}}&0.666 $\pm$ 0.05&0.720 $\pm$ 0.04&0.732 $\pm$ 0.04 \\
    Disabling A & 0.711 $\pm$ 0.04&0.584 $\pm$ 0.02&0.660 $\pm$ 0.03&0.676 $\pm$ 0.04 \\
    Disabling B & 0.742 $\pm$ 0.03&0.618 $\pm$ 0.13&0.695 $\pm$ 0.06&0.710 $\pm$ 0.06 \\
    Disabling C & 0.754 $\pm$ 0.04&\textbf{\underline{0.730 $\pm$ 0.06}}&\textbf{\underline{0.742 $\pm$ 0.05}}&\textbf{\underline{0.743 $\pm$ 0.04}} \\
    Disabling AB & 0.681 $\pm$ 0.02&0.334 $\pm$ 0.10&0.570 $\pm$ 0.03&0.599 $\pm$ 0.04 \\
    Disabling AC & 0.710 $\pm$ 0.02&0.606 $\pm$ 0.07&0.668 $\pm$ 0.04&0.678 $\pm$ 0.03 \\
    Disabling BC & 0.732 $\pm$ 0.04&0.630 $\pm$ 0.14&0.694 $\pm$ 0.07&0.705 $\pm$ 0.06 \\
    \hline
    \end{tabular}
    \caption{Results (Average $\pm$ SD) of Experiment 1: Yelp, CNNs} \label{tab:1a}
\end{table*}

\begin{table*}[t!] 
    \centering
    \small
    \begin{tabular}{|l|c|c|c|c|c|c|}
    \hline
    \multirow{2}{*}{\makecell[c]{Model: CNNs}} & \multicolumn{6}{c|}{\textbf{Test dataset: Amazon Products}} \\ \cline{2-7}
    & Clothes F1 & Music F1 & Office F1 & Toys F1 & Accuracy & Macro F1 \\ \hline
    Original&\textbf{\underline{0.806 $\pm$ 0.02}}&\textbf{0.960 $\pm$ 0.00}&0.789 $\pm$ 0.03&\textbf{\underline{0.748 $\pm$ 0.01}}&\textbf{\underline{0.825 $\pm$ 0.00}}&\textbf{\underline{0.829 $\pm$ 0.00}}\\
    Disabling A &0.724 $\pm$ 0.02&0.827 $\pm$ 0.06&0.722 $\pm$ 0.03&0.679 $\pm$ 0.03&0.738 $\pm$ 0.02&0.744 $\pm$ 0.02\\
    Disabling B &0.773 $\pm$ 0.02&0.956 $\pm$ 0.00&0.711 $\pm$ 0.02&0.688 $\pm$ 0.02&0.779 $\pm$ 0.02&0.785 $\pm$ 0.02\\
    Disabling C &0.786 $\pm$ 0.01&0.958 $\pm$ 0.01&\textbf{\underline{0.795 $\pm$ 0.02}}&0.734 $\pm$ 0.02&0.817 $\pm$ 0.00&0.821 $\pm$ 0.00\\
    Disabling AB &0.515 $\pm$ 0.08&0.586 $\pm$ 0.17&0.530 $\pm$ 0.04&0.512 $\pm$ 0.04&0.536 $\pm$ 0.05&0.556 $\pm$ 0.05\\
    Disabling AC &0.578 $\pm$ 0.11&0.745 $\pm$ 0.05&0.652 $\pm$ 0.04&0.579 $\pm$ 0.01&0.638 $\pm$ 0.03&0.669 $\pm$ 0.01\\
    Disabling BC &0.768 $\pm$ 0.02&0.948 $\pm$ 0.01&0.663 $\pm$ 0.06&0.627 $\pm$ 0.07&0.750 $\pm$ 0.04&0.754 $\pm$ 0.04\\
    \hline
    \end{tabular}
    \caption{Results (Average $\pm$ SD) of Experiment 1: Amazon Products, CNNs} \label{tab:1b}
\end{table*}

\begin{table*}[t!] 
    \centering
    \small
    \begin{tabular}{|l|c|c|c|c|}
    \hline
    \multirow{2}{*}{\makecell[c]{Model: BiLSTMs}} & \multicolumn{4}{c|}{\textbf{Test dataset: Yelp}} \\ \cline{2-5}
    & Negative F1 & Positive F1 & Accuracy & Macro F1 \\ \hline
    Original&\textbf{0.810 $\pm$ 0.01}&\textbf{0.774 $\pm$ 0.03}&\textbf{\underline{0.794 $\pm$ 0.01}}&\textbf{0.799 $\pm$ 0.01}\\
    Disabling A&\textbf{0.810 $\pm$ 0.00}&0.767 $\pm$ 0.01&0.791 $\pm$ 0.01&0.798 $\pm$ 0.00\\
    Disabling B&0.800 $\pm$ 0.00&0.745 $\pm$ 0.01&0.776 $\pm$ 0.01&0.785 $\pm$ 0.01\\
    Disabling C&0.803 $\pm$ 0.00&\textbf{0.774 $\pm$ 0.01}&0.790 $\pm$ 0.01&0.793 $\pm$ 0.00\\
    Disabling AB&0.781 $\pm$ 0.01&0.720 $\pm$ 0.02&0.754 $\pm$ 0.02&0.763 $\pm$ 0.02\\
    Disabling AC&0.800 $\pm$ 0.00&0.758 $\pm$ 0.01&0.781 $\pm$ 0.00&0.787 $\pm$ 0.00\\
    Disabling BC&0.787 $\pm$ 0.01&0.730 $\pm$ 0.02&0.762 $\pm$ 0.01&0.769 $\pm$ 0.01\\
    \hline
    \end{tabular}
    \caption{Extra results (Average $\pm$ SD) of Experiment 1: Yelp, BiLSTMs} \label{tab:1c}
\end{table*}

\begin{table*}[t!] 
    \centering
    \small
        \begin{tabular}{|l|c|c|c|c|c|c|}
    \hline
    \multirow{2}{*}{\makecell[c]{Model: BiLSTMs}} & \multicolumn{6}{c|}{\textbf{Test dataset: Amazon Products}} \\ \cline{2-7}
    & Clothes F1 & Music F1 & Office F1 & Toys F1 & Accuracy & Macro F1 \\ \hline
    Original&0.764 $\pm$ 0.01&\textbf{\underline{0.958 $\pm$ 0.00}}&\textbf{0.792 $\pm$ 0.02}&\textbf{0.760 $\pm$ 0.02}&\textbf{\underline{0.818 $\pm$ 0.01}}&\textbf{\underline{0.820 $\pm$ 0.01}}\\
    Disabling A&0.735 $\pm$ 0.03&0.940 $\pm$ 0.02&0.770 $\pm$ 0.02&0.733 $\pm$ 0.01&0.793 $\pm$ 0.01&0.796 $\pm$ 0.01\\
    Disabling B&0.747 $\pm$ 0.00&0.939 $\pm$ 0.02&0.765 $\pm$ 0.02&0.741 $\pm$ 0.01&0.798 $\pm$ 0.01&0.801 $\pm$ 0.01\\
    Disabling C&\textbf{\underline{0.769 $\pm$ 0.02}}&0.946 $\pm$ 0.01&\textbf{0.792 $\pm$ 0.03}&0.759 $\pm$ 0.04&0.816 $\pm$ 0.02&0.817 $\pm$ 0.02\\
    Disabling AB&0.636 $\pm$ 0.09&0.884 $\pm$ 0.04&0.720 $\pm$ 0.02&0.665 $\pm$ 0.04&0.727 $\pm$ 0.03&0.734 $\pm$ 0.02\\
    Disabling AC&0.718 $\pm$ 0.02&0.828 $\pm$ 0.08&0.758 $\pm$ 0.03&0.683 $\pm$ 0.03&0.745 $\pm$ 0.04&0.754 $\pm$ 0.04\\
    Disabling BC&0.702 $\pm$ 0.03&0.881 $\pm$ 0.05&0.702 $\pm$ 0.07&0.699 $\pm$ 0.03&0.750 $\pm$ 0.03&0.752 $\pm$ 0.03\\
    \hline
    \end{tabular}
    \caption{Extra results (Average $\pm$ SD) of Experiment 1: Amazon Products, BiLSTMs} \label{tab:1d}
\end{table*}

\clearpage

\begin{table*}[h!] 
    \centering
    \small
    \begin{tabular}{|l|c|c|c|c|c|c|}
    \hline
    \multirow{2}{*}{\makecell[c]{Model: CNNs}} & \multicolumn{6}{c|}{\textbf{Test dataset: Biosbias}} \\ \cline{2-7}
    & Surgeon F1 & Nurse F1 & Accuracy & Macro F1 & FPED $\downarrow$& FNED $\downarrow$\\ \hline
    Original &\textbf{\underline{0.957 $\pm$ 0.00}}&\textbf{\underline{0.943 $\pm$ 0.00}} & \textbf{\underline{0.951 $\pm$ 0.00}}&\textbf{\underline{0.950 $\pm$ 0.00}}&0.250 $\pm$ 0.02&0.338 $\pm$ 0.02 \\
    Disabling (MTurk) &0.943 $\pm$ 0.01&0.925 $\pm$ 0.01 & 0.935 $\pm$ 0.01&0.934 $\pm$ 0.01&0.163 $\pm$ 0.01&0.149 $\pm$ 0.03 \\ 
    Disabling (One) &0.942 $\pm$ 0.01&0.924 $\pm$ 0.01 & 0.934 $\pm$ 0.01&0.933 $\pm$ 0.01&\textbf{\underline{0.118 $\pm$ 0.00}}&\textbf{\underline{0.085 $\pm$ 0.01}} \\[2pt] \hline
    \end{tabular}
    \caption{Results (Average $\pm$ SD) of Experiment 2: Biosbias, CNNs} \label{tab:2b}
\end{table*}

\begin{table*}[h!] 
    \centering
    \small
    \begin{tabular}{|l|c|c|c|c|c|c|}
    \hline
    \multirow{2}{*}{\makecell[c]{Model: CNNs}} & \multicolumn{6}{c|}{\textbf{Test dataset: Waseem}} \\ \cline{2-7}
    & Not Abusive F1 & Abusive F1 & Accuracy & Macro F1 & FPED $\downarrow$ & FNED $\downarrow$ \\ \hline
    Original &\textbf{\underline{0.876 $\pm$ 0.00}}&\textbf{\underline{0.682 $\pm$ 0.01}} & \textbf{\underline{0.821 $\pm$ 0.00}}&\textbf{\underline{0.783 $\pm$ 0.00}}&0.232 $\pm$ 0.03&0.212 $\pm$ 0.02 \\
    Disabling (MTurk) &0.865 $\pm$ 0.00&0.671 $\pm$ 0.01 & 0.808 $\pm$ 0.00&0.770 $\pm$ 0.00&0.303 $\pm$ 0.02&0.220 $\pm$ 0.04 \\ 
    Disabling (One) &0.856 $\pm$ 0.01&0.614 $\pm$ 0.04 & 0.791 $\pm$ 0.02&0.743 $\pm$ 0.02&\textbf{0.205 $\pm$ 0.03}&\textbf{0.184 $\pm$ 0.03} \\[2pt] \hline
    \end{tabular}
    
    \vspace{4pt}
    
    \begin{tabular}{|l|c|c|c|c|c|c|}
    \hline
    \multirow{2}{*}{\makecell[c]{Model: CNNs}} & \multicolumn{6}{c|}{\textbf{Test dataset: Wikitoxic}} \\ \cline{2-7}
    & Not Abusive F1 & Abusive F1 & Accuracy & Macro F1 & FPED $\downarrow$ & FNED $\downarrow$ \\ \hline
    Original &\textbf{\underline{0.973 $\pm$ 0.00}}&0.179 $\pm$ 0.03 & \textbf{\underline{0.948 $\pm$ 0.00}}&0.601 $\pm$ 0.02&\textbf{0.052 $\pm$ 0.01}&0.164 $\pm$ 0.03 \\
    Disabling (MTurk) &0.967 $\pm$ 0.01&\textbf{\underline{0.230 $\pm$ 0.05}}& 0.936 $\pm$ 0.02&\textbf{0.609 $\pm$ 0.04}&0.083 $\pm$ 0.04&0.181 $\pm$ 0.05\\
    Disabling (One) &0.970 $\pm$ 0.00&0.191 $\pm$ 0.01 & 0.942 $\pm$ 0.01&0.598 $\pm$ 0.01&0.053 $\pm$ 0.00&\textbf{0.112 $\pm$ 0.02}\\[2pt]
    \hline
    \end{tabular}
    \caption{Results (Average $\pm$ SD) of Experiment 2: Waseem \& Wikitoxic, CNNs} \label{tab:2a}
\end{table*}

\begin{table*}[h!] 
    \centering
    \small
    \begin{tabular}{|l|c|c|c|c|}
    \hline
    \multirow{2}{*}{\makecell[c]{Model: CNNs}} & \multicolumn{4}{c|}{\textbf{Test dataset: 20Newsgroups}} \\ \cline{2-5}
    & Atheism F1 & Christian F1 & Accuracy & Macro F1 \\ \hline
    Original & \textbf{\underline{0.828 $\pm$ 0.01}}&\textbf{\underline{0.875 $\pm$ 0.01}}&\textbf{\underline{0.855 $\pm$ 0.01}}&\textbf{\underline{0.853 $\pm$ 0.01}} \\
    Disabling (MTurk) & 0.798 $\pm$ 0.01&0.853 $\pm$ 0.01&0.830 $\pm$ 0.01&0.828 $\pm$ 0.01 \\[2pt] \hline
    \end{tabular}
    
    \vspace{4pt}
    
    \begin{tabular}{|l|c|c|c|c|}
    \hline
    \multirow{2}{*}{\makecell[c]{Model: CNNs}} & \multicolumn{4}{c|}{\textbf{Test dataset: Religion}} \\ \cline{2-5}
    & Atheism F1 & Christian F1 & Accuracy & Macro F1 \\ \hline
    Original & 0.567 $\pm$ 0.03&0.787 $\pm$ 0.01&0.715 $\pm$ 0.02&0.731 $\pm$ 0.01 \\
    Disabling (MTurk) &\textbf{\underline{0.700 $\pm$ 0.15}}&\textbf{\underline{0.834 $\pm$ 0.04}}&\textbf{\underline{0.789 $\pm$ 0.07}}&\textbf{\underline{0.799 $\pm$ 0.06}}\\[2pt]
    \hline
    \end{tabular}
    \caption{Results (Average $\pm$ SD) of Experiment 3: 20Newsgroups \& Religion, CNNs} \label{tab:3a}
\end{table*}

\begin{table*}[h!] 
    \centering
    \small
    \begin{tabular}{|l|c|c|c|c|}
    \hline
    \multirow{2}{*}{\makecell[c]{Model: CNNs}} & \multicolumn{4}{c|}{\textbf{Test dataset: Amazon Clothes}} \\ \cline{2-5}
    & Negative F1 & Positive F1 & Accuracy & Macro F1 \\ \hline
    Original & \textbf{\underline{0.862 $\pm$ 0.01}}&\textbf{\underline{0.862 $\pm$ 0.01}}&\textbf{\underline{0.862 $\pm$ 0.01}}&\textbf{\underline{0.862 $\pm$ 0.01}}\\
    Disabling (MTurk) & 0.857 $\pm$ 0.01&0.855 $\pm$ 0.01&0.856 $\pm$ 0.01&0.856 $\pm$ 0.01 \\[2pt] \hline
    \end{tabular}
    
    \vspace{4pt}
    
    \begin{tabular}{|l|c|c|c|c|}
    \hline
    \multirow{2}{*}{\makecell[c]{Model: CNNs}} & \multicolumn{4}{c|}{\textbf{Test dataset: Amazon Music}} \\ \cline{2-5}
    & Negative F1 & Positive F1 & Accuracy & Macro F1 \\ \hline
    Original & 0.640 $\pm$ 0.02&\textbf{0.722 $\pm$ 0.01}&0.687 $\pm$ 0.01&0.695 $\pm$ 0.01\\
    Disabling (MTurk) & \textbf{\underline{0.668 $\pm$ 0.01}}&\textbf{0.722 $\pm$ 0.01}&\textbf{\underline{0.697 $\pm$ 0.01}}&\textbf{\underline{0.701 $\pm$ 0.01}} \\[2pt] \hline
    \end{tabular}
    
    \vspace{4pt}
    
    \begin{tabular}{|l|c|c|c|c|}
    \hline
    \multirow{2}{*}{\makecell[c]{Model: CNNs}} & \multicolumn{4}{c|}{\textbf{Test dataset: Amazon Mixed}} \\ \cline{2-5}
    & Negative F1 & Positive F1 & Accuracy & Macro F1 \\ \hline
    Original & 0.784 $\pm$ 0.01&0.799 $\pm$ 0.00&0.792 $\pm$ 0.01&0.793 $\pm$ 0.00\\
    Disabling (MTurk) & \textbf{\underline{0.793 $\pm$ 0.00}}&\textbf{\underline{0.801 $\pm$ 0.00}}&\textbf{\underline{0.797 $\pm$ 0.00}}&\textbf{\underline{0.797 $\pm$ 0.00}} \\[2pt] \hline
    \end{tabular}
    
    \vspace{4pt}
    
    \begin{tabular}{|l|c|c|c|c|}
    \hline
    \multirow{2}{*}{\makecell[c]{Model: CNNs}} & \multicolumn{4}{c|}{\textbf{Test dataset: Yelp}} \\ \cline{2-5}
    & Negative F1 & Positive F1 & Accuracy & Macro F1 \\ \hline
    Original & 0.767 $\pm$ 0.02&0.800 $\pm$ 0.00&0.785 $\pm$ 0.01&0.789 $\pm$ 0.01\\
    Disabling (MTurk) & \textbf{\underline{0.786 $\pm$ 0.00}}&\textbf{\underline{0.804 $\pm$ 0.00}}&\textbf{\underline{0.795 $\pm$ 0.00}}&\textbf{\underline{0.796 $\pm$ 0.00}} \\[2pt] \hline
    \end{tabular}
    \caption{Results (Average $\pm$ SD) of Experiment 3: Sentiment Analysis (Amazon Clothes), CNNs} \label{tab:3b}
\end{table*}

\end{document}